\documentclass[final]{cvpr}

\usepackage{times}
\usepackage{epsfig}
\usepackage{graphicx}
\usepackage{amsmath}
\usepackage{amssymb}
\usepackage{subcaption}

\usepackage[pagebackref=true,breaklinks=true,colorlinks,bookmarks=false]{hyperref}

\def\cvprPaperID{11003} 
\def\confYear{CVPR 2021}

\begin{document}

\title{CGAP2: Context and gap aware predictive pose framework for early detection of gestures}

\author{Nishant Bhattacharya \qquad Suresh Sundaram\\
{\tt\small bnishant@iisc.ac.in}, {\tt\small vssuresh@iisc.ac.in}\\ 
WIRIN Lab\\
Indian Institute of Science, Bengaluru, India\\

}

\maketitle

\begin{abstract}
With a growing interest in autonomous vehicles' operation, there is an equally increasing need for efficient anticipatory gesture recognition systems for human-vehicle interaction.  Existing gesture-recognition algorithms have been primarily restricted to historical data. In this paper, we propose a novel context and gap aware pose prediction framework(CGAP2), which predicts future pose data for anticipatory recognition of gestures in an online fashion. CGAP2 implements an encoder-decoder architecture paired with a pose prediction module to anticipate future frames followed by a shallow classifier. CGAP2 pose prediction module uses 3D convolutional layers and depends on the number of pose frames supplied, the time difference between each pose frame, and the number of predicted pose frames. The performance of CGAP2 is evaluated on the Human3.6M dataset with the MPJPE metric. For pose prediction of 15 frames in advance, an error of 79.0mm is achieved. The pose prediction module consists of only 26M parameters and can run at 50 FPS on the NVidia RTX Titan. Furthermore, the ablation study indicates supplying higher context information to the pose prediction module can be detrimental for anticipatory recognition. CGAP2 has a 1-second time advantage compared to other gesture recognition systems, which can be crucial for autonomous vehicles. 

\end{abstract}

\section{Introduction}

Self-driving cars have recently garnered a lot of attention from the research community due to their massive impact on our everyday lives. Companies like Tesla and Waymo have shown the use of real-time machine learning algorithms to solve complex problems and successfully derive a decision from a multitude of sensors. Autonomous driving is a set of numerous sub-problems that need to be individually solved to achieve complete self-driving and such solutions need to work on low power devices with very high efficiency. One such problem, which has failed to gain much traction, is gesture recognition for human interactions in a self-driving scenario. This form of gesture recognition poses some unique challenges compounded by the unavailability of appropriate datasets. 

The challenges for interactions between humans and autonomous vehicles are as follows. The first challenge occurs when gestures share a lot of the initial movement. For example, "Stop" and "Go" gestures have significant initial overlap, and systems may find it difficult to distinguish between them until late in the gesture life cycle. The second problem is background segmentation since the target subject may occur in various cluttered environments.
In case of autonomous vehicle, human interaction can happen in the form of traffic police or pedestrian/passenger. In general, while traffic police place themselves in the middle of an intersection, pedestrian/passengers occurrence is arbitrary. Gestures signaling oncoming traffic to slow down or stop can occur anywhere in the camera's field of view, which increases the complexity of detection. Perspective is the final challenge due to the natural limb occlusions that occur as a result of varying camera angles. As per our knowledge, it is the first time in the literature that such an end-to-end anticipatory gesture recognition is proposed to handle aforementioned challenges.

In this paper, we design an efficient end-to-end system called Context and Gap Aware Predictive Pose Framework (CGAP2), which tackles the anticipatory gesture recognition problems. CGAP2 reframes the standard gesture recognition problem and proposes a solution to resolve the ambiguity caused by similarities occurring early in the gesture life cycle. CGAP2 background segmentation is inspired by pose estimation, which is an effective way of extracting human subjects from any scene while also being a precursor to gesture recognition. While gesture recognition and pose prediction has been widely regarded as separate problems, we show that temporally correlating pose features is essential for recognizing actions. A set of frames (context) sampled at regular intervals (gap) controls the amount of relevant information being fed into the system, which is then used to predict future data, thus eliminating gesture ambiguity. This anticipatory behavior can also be deployed in an online setting, which hasn't been attempted before in the literature to the extent of our knowledge. Human3.6M \cite{h36m_pami} is chosen as the dataset since it supports multiple perspectives captured by four different cameras. Additionally, the dataset offers multiple subjects with different body structures moving within the field of view of the camera, necessitating motion tracking to be inherent to the model. CGAP2 achieves an error of $79.0$mm for pose prediction and a classification accuracy of $21\%$ which, to our knowledge, is the first classification baseline on the Human3.6M dataset. Furthermore, the ablation study helps in understanding the context-gap relationship on accuracy and time gap on anticipatory gesture recognition.

\section{Related Work}

\textbf{3D pose estimation.} 3D pose estimation has had a lot of attention from the research community and can be classified into three classes. The first class of algorithms rely on some form of multi-view approach to achieve very low errors on the MPJPE\cite{h36m_pami} metric. Notably, Iskakov et al.\cite{Iskakov_2019} use a multi-view approach relying on camera parameters to project feature maps onto volumetric representations which are then converted to 3D heatmaps. Multi-view approaches often rely on the limitations of the human skeleton and 3D geometry restrictions imposed by perspective changes which is exploited by Pavlakos et al.\cite{Pavlakos_2017}, Kocabas et al.\cite{DBLP:journals/corr/abs-1903-02330} and Rhodian et. al\cite{Rhodin_2018}. Heatmap outputs, while being an effective way to represent pose coordinates, require a maximum operation which is not differentiable and suffer from quantization error. Sun et al.\cite{Sun_2018} solve this problem by introducing an integral equation which jointly solves both heatmap representation and post-processed joint regression. The second class of algorithms extrapolate 3D coordinates from their 2D counterparts. Martinez et al.\cite{8237550} simply use extra linear regression layers with ReLU activations to achieve an average of 62.9mm of error on Human3.6M dataset. Chen et al.\cite{Chen_2016} suggest a probabilistic formulation of human 3D pose estimation problem and employ nearest neighbour algorithm to match candidate poses. Pavalkos et al. in\cite{Pavalkos_2016} discretize the output dimensions to voxels and use 2D poses for intermediate supervision in their networks.
Recently, there has been a surge in employing temporal information for improving pose estimations which form the last class of algorithms. Hossain and Little\cite{10.1007/978-3-030-01249-6_5} and Lin et al.\cite{line_2017} argue that while existing methods can accurately predict 3D poses from images, these poses are not temporally consistent. While \cite{10.1007/978-3-030-01249-6_5} uses LSTMs to reduce pose errors by 12\%, \cite{line_2017} uses recurrent pose machines to predict 3D from 2D poses and refine them across 3 stages.  

\textbf{Action recognition.} Action recognition, gesture recognition and video classification are essentially synonyms for the same underlying task with subtle differences in the target demographics. Due to the temporal nature of datasets, most approaches use some form of temporal modelling either in the form of LSTMs, 3D Convolutions or Convolutional LSTMs. For instance, Zhang et al.\cite{ZhuNIPS2018} introduce attention based convolutional LSTMs and ascertain how they barely contribute to spatio-temporal feature fusion based on which they derive a new LSTM cell. Similarly, LRCNs\cite{Donahue_2017} use convolutions and LSTMs separately to learn variable length video representations across time. However, 3D convolutions have been proven to be efficient for comprehending spatio-temporal data which is shown in \cite{Kalfaoglu2020LateTM}, \cite{bourdev_2015} and \cite{sun_2015}. While C3D\cite{bourdev_2015} simply stacks linear layers after 3D Convolutions, BERT based temporal modelling in \cite{Kalfaoglu2020LateTM} incorporates an attention based bidirectional encoding mechanism. F\textsubscript{ST}CN\cite{sun_2015} factorizes a conventional 3D convolutional kernel into its  constituent 2D spatial kernels and 1D temporal kernels thereby reducing operational complexity. We also see another class of algorithms which tend to fuse feature maps from different streams. Feichtenhofer et al.\cite{Feichtenhofer_2016} develop separate temporal and spatial representations which are then fused later in the pipeline. Motion fused frames\cite{Kopuklu_2018_CVPR_Workshops} merge motion information with spatial information for hand gesture recognition. Tang et al.\cite{tang2019fast} introduce efficiency by removing images with low entropy and only extracting consequential frames. Data augmentation algorithms like \cite{zhang2019pan} rely on finding alternatives to optical flow by using persistence of appearance method. Temporal shift modules in \cite{lin2019tsm} instead shift the intermediate feature maps to effectively merge the multiple and add operations seen in convolutional layers.

\begin{figure*}
\begin{center}
\includegraphics[width=\linewidth]{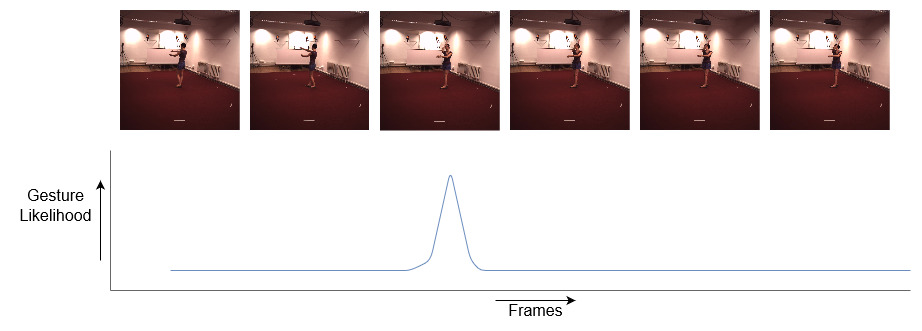}
\end{center}
   \caption{Gesture progression across video frames}
\label{fig:gesture_curve}
\end{figure*}

\textbf{Video prediction.} While action recognition and video prediction share a lot of the temporal challenges there are drastic differences in the approach due to differing outcomes. While action recognition concerns itself only with final class values, video prediction demands pixel wise accuracy. Generative adversarial approaches like \cite{kumar_2020} and \cite{aigner_2019} use auto encoders to directly predict pixels of the next frame, albeit with different philosophies. Finn et al.\cite{finn_2016} predict a distribution over pixel motion which partially mitigates complications introduced by perspective changes. Minderer et al.\cite{minderer2019unsupervised} reduce the complexity of pose prediction problem by using keypoints of an image as an effective distillation of spatial data. Zhu et al. \cite{zhu_2019} use video prediction to scale up existing datasets by jointly predicting future frames and their corresponding labels. U-net\cite{liu_2018} reframes the problem of video prediction as anomaly detection and argues that conventional reconstruction errors cannot tackle abnormal events. Wang et al.\cite{wang2018eidetic} and Byeon et al.\cite{byeon_2018} formulate their own versions of LSTM. While \cite{wang2018eidetic} merges 3D convolutions and recurrent neural networks to create eidetic 3D LSTMs, \cite{byeon_2018} introduces parallels multi-dimensional LSTMs which captures the entire past context by using blending units as their aggregators.  Oliu et al.\cite{oliu_2018} uses bijective GRUs which treats input as another recurrent state and constructs an autoencoder network by stacking multiple bGRUs. Hsieh et al. \cite{hsieh2018learning} decompose an image into its individual moving components and predict their paths separately. \cite{ye2019cvp} also uses a similar strategy and uses an scene's constituent objects to create a global trajectory.

\section{CGAP2 Framework}

Each gesture can be perceived as a sequence of frames where each frame corresponds to a likelihood metric indicating how representative it is of that gesture. One could view the likelihood metric as a Gaussian curve superimposed on the sequence where the peak of the curve corresponds to the frames which have the highest likelihood of getting classified as the target gesture as represented in Figure \ref{fig:gesture_curve}. In this figure, a typical gesture sequence for the class "Directions" is shown and it is observed that the peak corresponds to a drastic change in the initial sequence thereby demarcating its class. From the perspective of a gesture recognition system, the initial frames of all gestures are ambiguous, and it is unable to distinguish between classes until the frames corresponding to peaks are reached.  We maintain that gestures, specifically, traffic gestures share similar likelihood curves although the peaks might occur at different times depending on the class. This property is exploited to anticipate the peak of a gesture and to classify it even before the gesture has occurred, a feature crucial for self-driving applications. In this paper, we present a context and gap aware pose prediction (CGAP2) framework for anticipatory gesture recognition. The pose prediction module and anticipatory gesture recognition are two important components in CGAP2 and are described in the following subsections. 

\subsection{Context and gap aware pose prediction framework}
The problem statement is formulated mathematically as follows. Given a set of images $I_t$ where $t \in \{j, \cdots, j + (n \times g)\}$ CGAP2 network $F$ outputs future pose frames $P_t\prime$ where $t' \in \{1, \cdots,k \}$.
\begin{equation}
 F\left(I_t\right)=P_t\prime
 \label{eq:formulation}
\end{equation}
In equation \ref{eq:formulation} pose frames $P$ are poses corresponding to each frame and  $j$ is the starting sampling position in the video sequence. While ideally it should be somewhere before the peak of the gesture curve, one can see that it is arbitrarily assigned by the dataset. Here, $n$ and $g$ are defined as the context and gap variables respectively. Context dictates how much historical data is being fed to $F$. The consecutive frames have high redundancy of information which can lead to repetitive information being fed into our system. One can reduce this redundancy by skipping frames in between timestamps. Finally, k-value decides the required number of predicted pose frames. For this work, the context, gap and k-value are set as $5$, $15$ and $1$ respectively. In performance evaluation section, an ablation study is conducted to understand the effect of these parameters on anticipatory gesture recognition.

\begin{figure*}
     \centering
     \begin{subfigure}[b]{\textwidth}
         \centering
         \includegraphics[width=\textwidth]{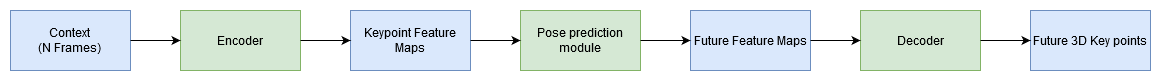}
         \caption{CGAP2 pose prediction module}
         \label{fig:pp_basic_diagram}
     \end{subfigure}
     \hfill
     \begin{subfigure}[b]{\textwidth}
         \centering
         \includegraphics[width=\textwidth]{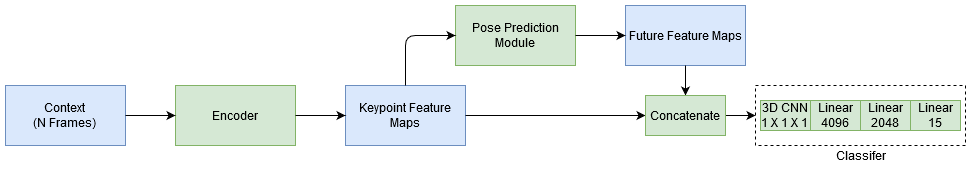}
         \caption{CGAP2 anticipatory classifier using pre-trained pose prediction module}
         \label{fig:classification}
     \end{subfigure}
        \caption{Schematic diagram for CGAP2 framework. The blue boxes denote data while the green boxes denote operations performed on the data.}
        \label{fig:architectures}
\end{figure*}


\begin{figure*}
\begin{center}
\includegraphics[width=\linewidth]{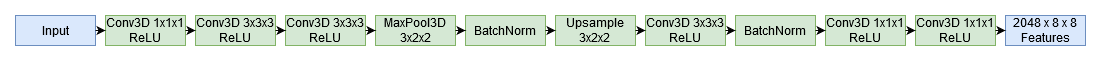}
\end{center}
   \caption{Architecture of deep neural pose prediction module}
\label{fig:pp_temporal}
\end{figure*}

Figure \ref{fig:pp_basic_diagram} gives a high-level overview of the pose prediction framework. We take inspiration from Kocabas et al. \cite{DBLP:journals/corr/abs-1903-02330} and create a  encoder-decoder architecture with the temporal module inserted in between for post prediction module. Conceptually, instead of directly predicting future pose frames CGAP2 predicts future feature maps and then passes it through through the decoder.  The motivation behind doing this is two fold, the first one being that while the decoder outputs heatmaps of $64 \times 64 \times 64$ for $17$ joints, the encoded features are only $8 \times 8 \times 2048$. An encoder module is incorporated to handle the difference in dimensionality. Secondly, CGAP2 approach is generalisable over any task which derives temporal coherency from existing precursor algorithms. 

The encoder and decoder use pre-trained weights from Kocabas et al. \cite{DBLP:journals/corr/abs-1903-02330}. It is ensured that the temporal module only establishes the relationship across time instances since the other two components only contain 2D convolutions and therefore are restricted only to spatial comprehension. For the temporal module, we use 3D convolutions since they have been known to be effective for time series data. Convolutional LSTMs on the other hand have a dependency on loops which exponentially increases the number of operation during run-time according to our experiments. 

The architecture of deep pose prediction model is shown in Figure \ref{fig:pp_temporal}. It consists of a bottleneck layer of $1 \times 1 \times 1$ filters which reduces the dimensionality from $2048$ to $512$ channels followed by two more convolutional layers each with $3 \times 3 \times 3$ filters. Max pooling and batch normalization layers are introduced to reduce over fitting in the network. However, to avoid information loss, up sampling layers followed by two convolutional layers are used. Each convolutional layer is normalized using ReLU activation. The number of layers have been restricted to limit the amount of overhead caused by the pose prediction module. Furthermore, we note that the decoder is only required for the training of pose prediction module and is not required for the final anticipatory classification. L1 distance between the predicted future pose and the target future pose over 17 joints is taken as the loss function as represented in equation \ref{eq:l1}.

\begin{equation}
 L = \sum_{i=1}^{17} \lvert y_{target} - y_{predicted} \rvert
 \label{eq:l1}
\end{equation}

The architecture described here is designed to be small with a total of $50$ million parameters with only $26$ million parameters for the pose prediction module. It can run at $~50$ FPS on an NVidia RTX titan and $~30$ FPS on an NVidia 960M.

\subsection{Anticipatory Gesture classification}
Anticipatory gesture classification tasks follow a generic blueprint which consists of feature construction phase followed by a classification phase. CGAP2 hypothesizes that most of the feature map construction has been done by the encoder. In figure \ref{fig:classification} one can see that pose prediction module trained in the previous phase has been used and the predicted feature maps are concatenated with the encoder feature maps before being fed into the classifier. This enables our classifier to perceive both historical and future data to make a classification. As mentioned previously, for classification the decoder module in pose prediction is ignored. However, since a classification task needs a different feature map distillation, few 3D convolutional layers are introduced before the final fully connected layers to convert the pose feature maps to anticipatory classification feature maps 



The CGAP2 anticipatory gesture classifier consists of a single 3D convolutional layer which is both responsible for reducing the dimensionality and morphing the pose feature maps to the required anticipatory classification feature maps. Efficiency becomes even more important since the output of the convolutions is fed into fully connected layers which often have the highest number of operations. We reduce the feature maps to 4096 and 2048 feature points subsequently before the finally classification layer. Categorical cross entropy is used as the loss function during training. In the following sections, performance of both the pose prediction module and classification module is gauged.

\section{Performance evaluation}
In this section the choice of dataset is described followed by hyperparameter documentation in the training section for both the pose prediction and classifier modules. While quantitative results are also presented for both modules, the qualitative results only highlight the future pose frames predicted by CGAP2. Finally, an ablation study is conducted for variances in context, gap and architecture. The study also establishes a trade-off between pose prediction accuracy and time advantage for anticipatory path planning algorithms.

\subsection{Dataset}
A subset of the Human 3.6M \cite{h36m_pami} dataset called the Human50K dataset is used for training. It has a much smaller sample space with a 80-20 split between training and validation sets respectively with an image size of $256 \times 256$. Furthermore, the 3D coordinates of each pose is projected onto image space using the camera transformation matrix. As stated before, this dataset has been randomly sampled from its super set and hence we have no control over $j$.

\subsection{Training procedure}
Phase-wise training has been proven to be advantageous for a multitude of tasks and we employ the same here to train the pose prediction module. We first use the architecture shown in Figure \ref{fig:pp_basic_diagram} to train the pose prediction model and finally train anticipatory gesture recognition module \ref{fig:classification}.

{\bf Pose prediction module training.} The encoder and decoder are frozen and only the pose prediction module is trained in this phase. Stochastic gradient descent\cite{kiefer1952} is used with a momentum of 0.9 and a learning rate of 0.001 for 15 epochs. A weight decay of 0.0001 and a change in the learning by a factor 0.1 is applied after 5 epochs. The batch size is set to 32 and L1 distance is used as the loss function. CGAP2 takes around 30 minutes to train on a NVidia RTX Titan.

{\bf Classification module training.} Since the decoder is not required for this phase it is and the encoder and the pose prediction module is frozen. The same configuration with the same values for momentum, learning rate, epochs and weight decay is used. However, the batch size is increased to 64 and categorical cross entropy is used as the loss function. The time taken for each epoch is roughly the same as in phase 1.

\subsection{Quantitative results}
The pose prediction module performance is reported using Mean Per Joint Position Error(MPJPE)\cite{h36m_pami}. As per our knowledge, there aren't any models which demand a direct comparison with CGAP2 since it predicts pose frames in advance. However, we compare our model with the current and previous state-of-the-art approaches to establish a performance benchmark in Table \ref{tab:mpjpe}.

\setlength{\tabcolsep}{1pt}
\begin{table*}
\begin{center}
\begin{tabular}{|l c c c c c c c c c c c c c c c c|}
\hline
Method & Dir. & Disc. & Eat. & Greet. & Phone. & Photo & Pos. & Purch. & Sit. & SitD. & Smoke & Wait. & WalkD. & Walk. & WalkT. & Avg\\
\hline
Pavlakos et al.\cite{Pavlakos_2017} & 41.1 & 49.1 & 42.7 & 43.4 & 55.6 & 46.9 & 40.3 & 63.6 & - & - & - & - & - & - & - & -\\
Rogez et al.\cite{Rogez_2019} & 50.9 & 55.9 & 63.3 & 56.0 & 65.1 & 70.7 & 52.1 & 51.9 & 81.1 & 91.7 & 64.7 & 54.6 & 44.7 & 61.1 & 53.7 & 61.2\\
Martinez et al.\cite{8237550} & 51.8 & 56.2 & 58.1 & 59.0 & 69.5 & 78.4 & 55.2 & 58.1 & 74.0 & 94.6 &  62.3 & 59.1 & 65.1 & 49.5 & 52.4 & 62.9 \\
Sun et al.\cite{Sun_2018} & - & - & - & - & - & - & - & - & - & - & - & - & - & - & - & 49.6\\
Pavllo et al.\cite{Pavllo_2019} & 45.2 & 46.7 & 43.3 & 45.6 & 48.1 & 55.1 & 44.6 & 44.3 & 57.3 & 65.8 & 47.1 & 44.0 & 49.0 & 32.8 & 33.9 & 46.8 \\
Hossain et al.\cite{10.1007/978-3-030-01249-6_5} & 48.4 & 50.7 & 57.2 & 55.2 & 63.1 & 72.6 & 53.0 & 51.7 & 66.1 & 80.9 & 59.0 & 57.3 & 62.4 & 46.6 & 49.6 & 58.3 \\
Iskakov et al.(Mono)\cite{Iskakov_2019} & 41.9 & 49.2 & 46.9 & 47.6 & 50.7 & 57.9 & 41.2 & 50.9 & 57.3 & 74.9 & 48.6 & 44.3 & 41.3 & 52.8 & 42.7 & 49.9 \\
Iskakov et al.(Multi)\cite{Iskakov_2019} & \textbf{19.9} & \textbf{20.0} & \textbf{18.9} & \textbf{18.5} & \textbf{20.5} & \textbf{19.4} & \textbf{18.4} & \textbf{22.1} & \textbf{22.5} & \textbf{28.7} & \textbf{21.2} & \textbf{20.8} & \textbf{19.7} & \textbf{22.1} & \textbf{20.2} & \textbf{20.8} \\

\hline
\hline
Ours & 59.2 & 68.8 & 74.6 & 81.0 & 71.4 & 79.7 & 62.9 & 69.8 & 59.8 & 75.3 & 64.3 & 69.9 & 101.1 & 156.6 & 133.8 & 79.0\\
\hline
\end{tabular}
\end{center}
\caption{Results are reported on the Human3.6M dataset with the metric being MPJPE in millimeters. Note that we have included both monocular and multi view approaches.}
\label{tab:mpjpe}
\end{table*}
\setlength{\tabcolsep}{6pt}

\begin{figure}
     \centering
     \begin{subfigure}[b]{0.46\textwidth}
         \centering
         \includegraphics[width=\textwidth]{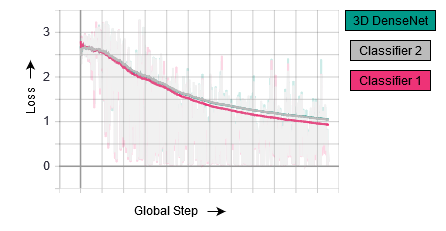}
         \caption{Training loss}
         \label{fig:classifier_loss}
     \end{subfigure}
     \hfill
     \begin{subfigure}[b]{0.46\textwidth}
         \centering
         \includegraphics[width=\textwidth]{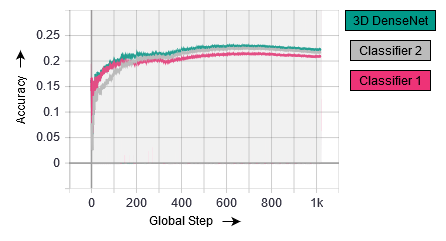}
         \caption{Validation accuracy}
         \label{fig:classifier_accuracy}
     \end{subfigure}
        \caption{CGAP2 Classification statistics.}
        \label{fig:classification_stats}
\end{figure}

Furthermore loss and accuracy of the classifier are stated in Figure \ref{fig:classification_stats}. 3 classifiers were tested including our own classifier and two versions of 3D DenseNet\cite{huang2019convolutional} with 32 and 100 layers respectively. CGAP2 classifier achieves an accuracy of 21\%\ and reports negligible improvements over 3D DenseNet in either loss or accuracy by increasing the number of layers. This can be attributed to the value of $j$ and its arbitrary nature in the dataset. Sampling frames just before the curve for each gesture will improve the classifier accuracy drastically. 



\subsection{Qualitative results}
\begin{figure}
   \centering
\begin{tabular}{|l c c|}
\hline
Output Frame & GT Pose Plot & Predicted Pose Plot\\
\hline
\includegraphics[width=0.28\linewidth]{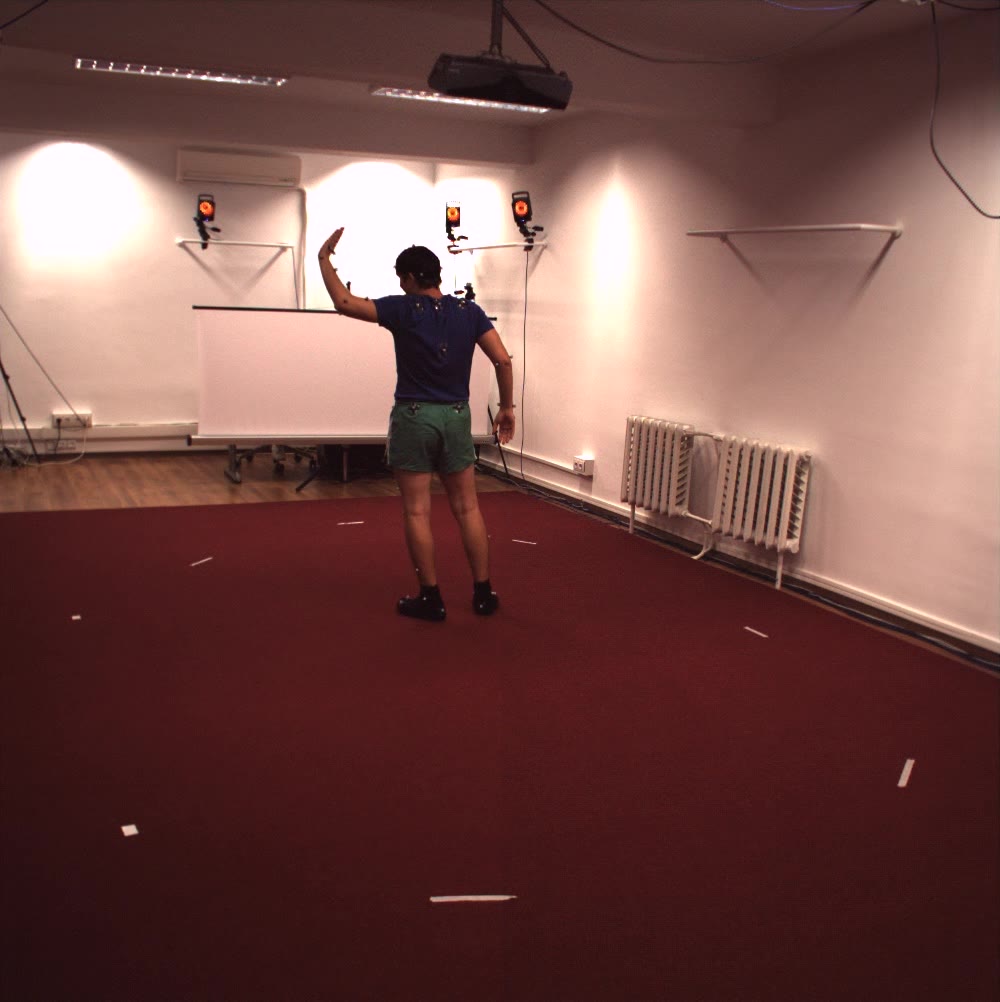} & \includegraphics[width=0.3\linewidth]{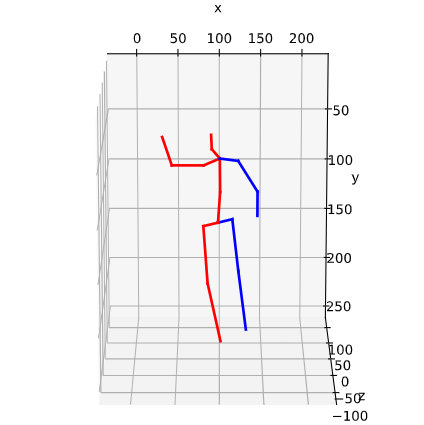} & \includegraphics[width=0.3\linewidth]{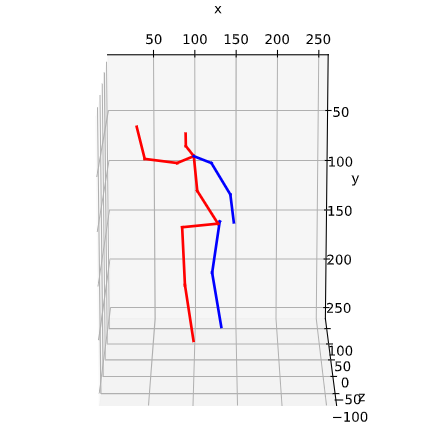}\\
\hline
\includegraphics[width=0.28\linewidth]{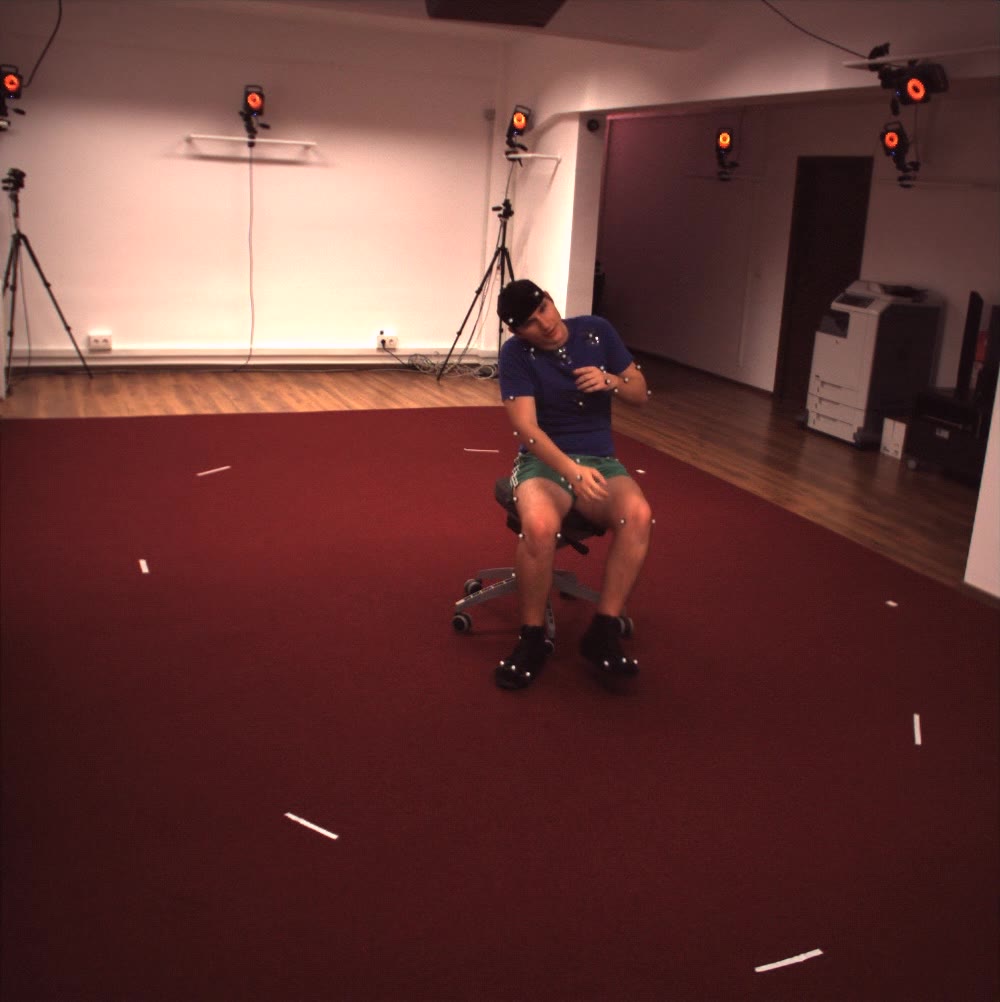} & \includegraphics[width=0.3\linewidth]{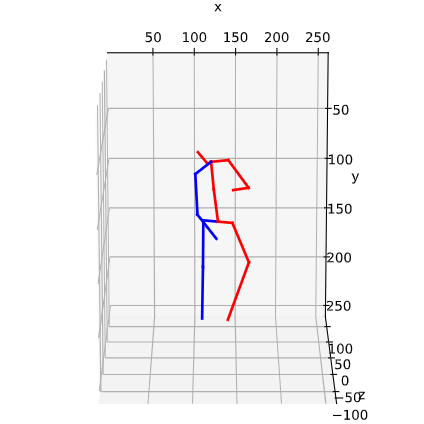} & \includegraphics[width=0.3\linewidth]{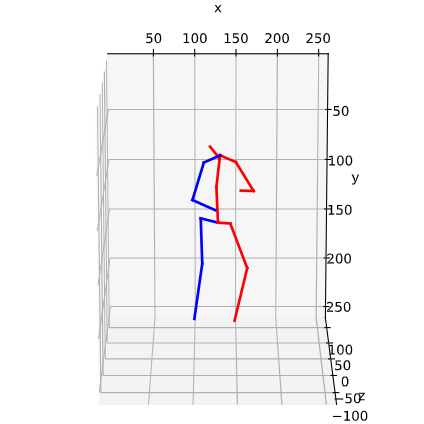}\\
\hline
\includegraphics[width=0.28\linewidth]{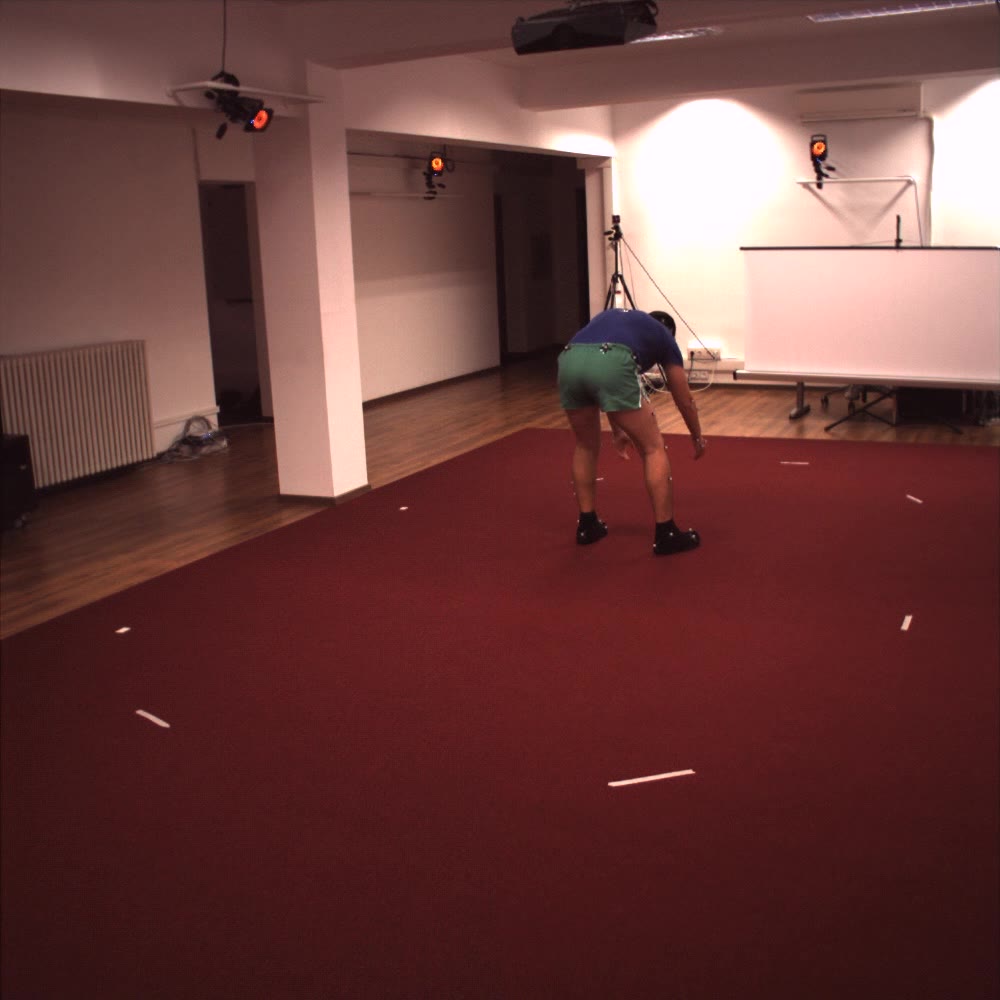} & \includegraphics[width=0.3\linewidth]{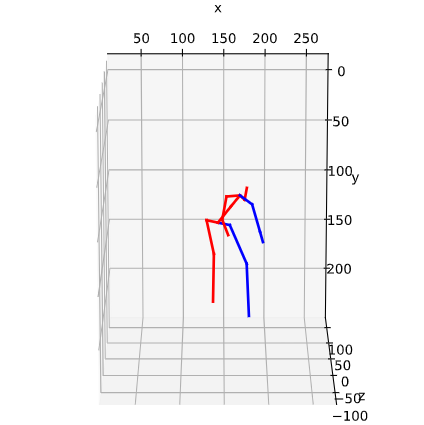} & \includegraphics[width=0.3\linewidth]{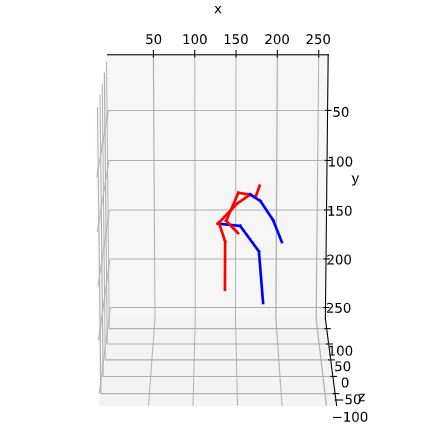}\\
\hline
\end{tabular}

    \caption{Qualitative results on the Human3.6M dataset}
    \label{fig:qual_pose} 
\end{figure}

In this sub section some qualitative results of our pose prediction module are presented in Figure \ref{fig:qual_pose}. The left most column represents the (i + g)\textsuperscript{th} where $i$ represents the last image in the context and $g$ is gap with a value of 15. The middle and the last columns represent the ground truth and predicted labels respectively. It is observed that while the pose predicting module is able to reasonably predict the future poses it suffers from some inconsistencies in the pose structure. This behaviour is attributed to the inherent temporal-only design of the pose prediction module which inhibits any form of human skeleton reasoning.

\subsection{Ablation study}
\begin{figure}
     \centering
     \begin{subfigure}[b]{0.46\textwidth}
         \centering
         \includegraphics[width=\textwidth]{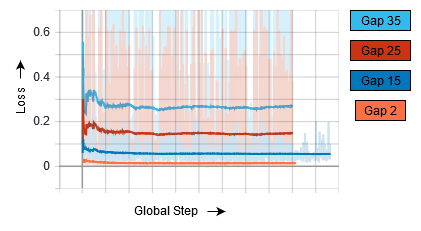}
         \caption{Gap variation}
         \label{fig:gap_variation}
     \end{subfigure}
     \hfill
     \begin{subfigure}[b]{0.46\textwidth}
         \centering
         \includegraphics[width=\textwidth]{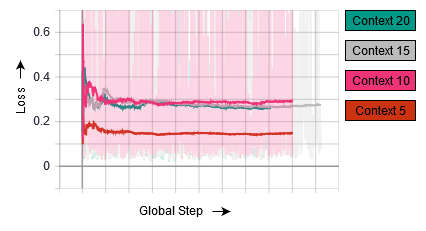}
         \caption{Context variation}
         \label{fig:context_variation}
     \end{subfigure}
     \hfill
     \begin{subfigure}[b]{0.46\textwidth}
         \centering
         \includegraphics[width=\textwidth]{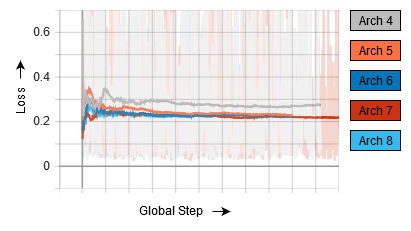}
         \caption{Architecture variation}
         \label{fig:arch_variation}
     \end{subfigure}
     \hfill
     \begin{subfigure}[b]{0.46\textwidth}
         \centering
         \includegraphics[width=\textwidth]{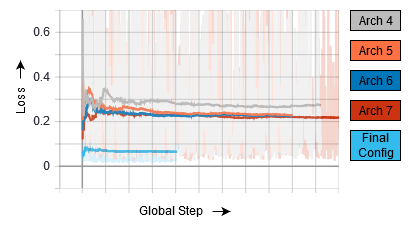}
         \caption{Final architecture variation}
         \label{fig:final}
     \end{subfigure}
        \caption{Ablation testing. Context, gap and architectures are varied to find the permutation which performs the best.}
        \label{fig:ablation_testing}
\end{figure}

In this section we justify the context, gap and architecture choices with the help of experiments conducted on the Human50K dataset. Each of these experiments were conducted with the same configuration as stated in the training section. However, to lend credibility the number of epochs was increased to 50. The y-axis of each graph is the validation loss reported for each global step in the validation loop which is reported on the x-axis. 

We first test to ascertain the ideal values for gap by conducting experiments at values of 2, 15, 25 and 35 in Figure \ref{fig:gap_variation}. While technically a gap level of 2 gives the best results, it is impractical for efficiency. If one considers a camera recording at 15 FPS, selecting a gap of 15 will give CGAP2 a 1 second advantage over any online classification algorithm employing historical data. In contrast a gap of 2 reduces this advantage to a mere 0.13 seconds. This is crucial in a self-driving scenario since being able to predict events 1 second in advance leaves more room for decision making.

For context experimentation, the gap is arbitrarily selected as 25 and the architecture is frozen with only a single 3D convolutional layer. From the Figure \ref{fig:context_variation}  one can see that a context of 5 outperforms 10, 15 and 20. The two peculiar behaviours to note is that adding more context leads to a worse result which is opposite of our initial intuition and adding even more context afterwards has a negligible effect. This behaviour can be explained by the hypothesis that owing to a large gap value, context can quickly become irrelevant or even detrimental to the final pose prediction due to information loss. This hypothesis also explains the second anomaly we observe since adding even more information can be seen as adding further inconsequential data which has no affect on the results. It is, therefore, concluded that a context value of 5 is ideal.

Finally, experiments are carried out with different architectures till the point of diminishing returns is reached. We start with a single 3D convolution layer gradually adding more of the same along with residual connections, batch normalization and max pooling. We experiment with 5 different architectures all with a gap and context of 15 each. It is observed that there isn't much of an improvement in the results although the network converges faster in some cases. It is concluded that adding more layers would increase the computational complexity. Final architecture for the pose prediction module is defined in Figure \ref{fig:pp_temporal}. We demonstrate the improvements in Figure \ref{fig:final} but we limit the final run to only 15 epochs since the later epochs don't result in any significant improvements.

\section{Conclusion and Future Work}
In this paper, we have presented a novel context and gap aware pose prediction framework for anticipatory recognition of the gesture in an online fashion. CGAP2 implements an encoder-decoder architecture paired with a pose prediction module to anticipate future frames followed by a shallow classifier. CGAP2 performance is evaluated using the Human3.6M dataset with the MPJPE metric. The results clearly indicate that the CGAP2 achieves an error of 79.0mm with 15 frames in advance. The ablation study suggests the trade-off between time advantage and accuracy and has a one-second time advantage over the other gesture recognition system. Future work focuses on improving anticipatory gesture recognition accuracy and reduces the structure of the pose prediction model. Further, one needs to optimize the encoder-decoder model to enhance the anticipatory pose, essential for accurate recognition.

{\small
\bibliographystyle{ieee_fullname}
\bibliography{egbib}
}

\end{document}